\documentclass[runningheads,a4paper]{llncs}

\usepackage{graphicx}
\usepackage{layout}
\usepackage{amssymb}
\usepackage{amsmath}
\usepackage{graphicx}
\usepackage{fancyhdr}
\usepackage{epigraph}

\usepackage{subfigure}
\usepackage{varwidth,xcolor}

\usepackage{url}   
\newcommand{\keywords}[1]{\par\addvspace\baselineskip
\noindent\keywordname\enspace\ignorespaces#1}

\newtheorem{defn}{Definition}

\begin{document}

\mainmatter  

\title{STAM: A Framework for Spatio-Temporal Affordance Maps}

\titlerunning{STAM: A Framework for Spatio-Temporal Affordance Maps}

%
%
\author{Francesco Riccio\inst{1} \and Roberto Capobianco\inst{1} \and
  Marc Hanheide\inst{2} \and Daniele Nardi\inst{1}}
\authorrunning{STAM: A Framework for Spatio-Temporal Affordance Maps}

\institute{
  Department of Computer, Control, and Management Engineering\\
  Sapienza University of Rome\\
  via Ariosto 25, Rome, 00185 Italy\\
  \email{\{riccio, capobianco, nardi\}@dis.uniroma1.it} \and
  Lincoln Centre for Autonomous Systems, School of Computer Science\\
  University of Lincoln\\
  Brayford Pool, Lincoln, Lincolnshire, LN6 7TS, United Kingdom\\
  \email{mhanheide@lincoln.ac.uk}
}

%
%

\toctitle{STAM: A Framework for Spatio-Temporal Affordance Maps}
\maketitle

\begin{abstract}
  Affordances have been introduced in literature as action
  opportunities that objects offer, and used in robotics to
  semantically represent their interconnection. However, when
  considering an environment instead of an object, the problem becomes
  more complex due to the dynamism of its state. To tackle this
  issue, we introduce the concept of Spatio-Temporal Affordances
  (STA) and Spatio-Temporal Affordance Map (STAM). Using this
  formalism, we encode action semantics related to the environment to
  improve task execution capabilities of an autonomous robot. We
  experimentally validate our approach to support the execution 
  of robot tasks by showing that affordances encode accurate 
  semantics of the environment.
\end{abstract}

\keywords{Spatial Knowledge, Affordances, Semantic Agents}

\section{Introduction}
\label{sec:intro}

The concept of affordances has been originally introduced by
Gibson~\cite{Gibson1979} as action opportunities that objects
offer. This idea has been recently used in robotics to
learn~\cite{Koppula2013}, represent~\cite{Pandey2012} and
exploit~\cite{Kim2015} object related actions in human-populated
environments. However, when considering the affordances of an
environment, methods proposed in literature cannot be directly
applied. Differently from normal objects, the state of the environment
is highly dynamic and contains the state of the robot and other
dynamic entities, such as humans. This inevitably leads to a more
complex problem that requires specific representation and learning
approaches.

To tackle this problem, the concept of spatial affordance has been
adopted in some works with the aim of supporting
navigation~\cite{Epstein2015} or improving the performance of a
tracking system. In this work, we use this concept to encode action
semantics related to the environment to improve task execution
capabilities of an autonomous robot. In particular, we formalize a
Spatio-Temporal Affordance Map (STAM) as a representation that
contains high-level semantic properties of an environment, directly
grounded on the operational scenario. This grounding is obtained
through the use of a function (the affordance function), that
generates areas of the environment that afford an action, given a
particular state or an equivalent observation of the world. More in
detail, STAM contains generic descriptors that (if needed) provide
prior information about the actions. For example, when performing a
following task, we might not want the relative distance of a robot,
with respect to the followed individual, to be greater than a given
threshold. These descriptors are then specialized according to the
environment where the robot is operating -- i.e., the current state of
the external world, its entities, including objects and people, and
their position over time.

\begin{figure}[t!]
  \centering
  \includegraphics[width=\textwidth]{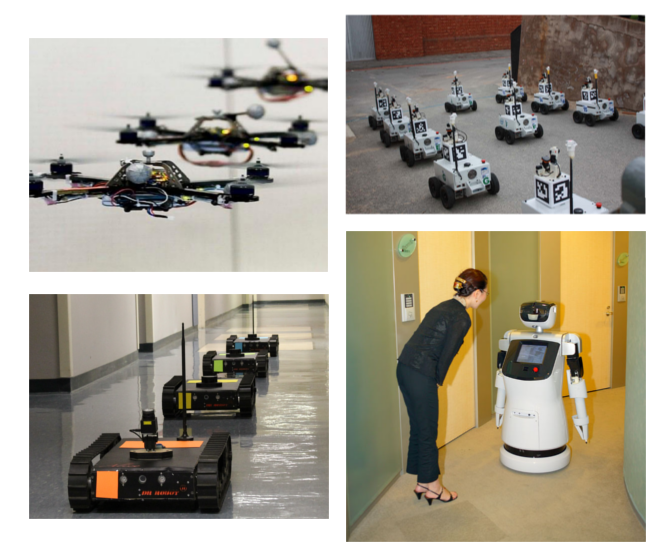}
  \caption{Following is a key skill in several robotic applications:
    swarm airdrones, robot teaming, exploration and service tasks. All
    of them, however, require the robot to execute the task according
    to different criteria, such as closeness or social
    acceptability. Being able to represent (and learn) the task
    semantics according to the specific scenario improves task
    execution capabilities of a robot.}
  \label{fig:intro}
\end{figure}

We evaluate an autonomous STAM agent over the execution of a following
task that, as shown in \figurename~\ref{fig:intro}, can be beneficial
in several applications. In this example, we use expert demonstrations
to teach a robot the spatial relation that holds between the
environment and the task ``to follow''. While learning from
demonstration has been already used to learn object
affordances~\cite{Montesano2008}, we provide an example of how to
easily extend such techniques to the case of spatial affordances.  Our
tests demonstrate that affordances encode accurate semantics of the
environment and that they can be used to improve robot skills in terms
of efficiency and acceptability in the specific context.

The remainder of this paper is organized as
follows. Section~\ref{sec:related} presents previous research about
affordances, while Section~\ref{sec:stam} defines the concept of
Spatio-Temporal Affordances and Spatio-Temporal Affordance Maps, by 
also describing how they can be generated
(Section~\ref{sec:generating_stam}). Additionally,
Section~\ref{sec:stam_robot} describes how to use STAM on a robot and
Section~\ref{sec:experiments} reports on our experimental
validation. Final conclusions are presented in
Section~\ref{sec:conclusion}.

\section{Related Work}
\label{sec:related}
Affordance theory has been introduced to represent possible actions
that a robot can perform over a particular object. We extend
affordance theory to explicitly formalize the environment itself as a
combination of spatial affordances that are used to provide a semantic
analysis of the space surrounding the robot. In this context, there
is not a vast literature that represent spatial affordances
and no prior work models affordances in a general
framework. In fact, affordances are used to leverage a particular
robot behavior or to adapt the routine of a specific algorithm. For
example, Epstein et al.~\cite{Epstein2015} exploit spatial affordances
to support navigation. In this work, the leaned spatial affordance
informs the robot about the most suitable action to execute for
navigation. However, this approach cannot be generalized, since the
affordance model strictly depends on a metric representation of the
operational scenario. Hence, different representations, such as topological maps, cannot be used. 
Similarly to our work, Diego et al.~\cite{Diego2011} encode activities in an affordance map in order to leverage robot movements. The affordance map is used to represent the presence
of people in the environment and then to avoid crowded areas not easily navigable. In a different
scenario, Luber et al.~\cite{Luber2011} use affordances to
improve tracking and prediction of people destinations. Also in this
case, the authors exploit spatial affordances to map activities
directly into the operational scenario. However, their system is not
intended to run on a robot, and the activities recognized only relate
to the presence of people in the scenario. The aforementioned works
formalize spatial affordances to only represent navigability of the
environment, and in most of the cases, the proposed approaches cannot
encode spatial semantics which is a key contribution of our work.

Manifold works confirm our insights that a proper spatial
semantic representation can improve robot capabilities. These works typically
evaluate spatial semantics although they do not explicitly
represent spatial affordances. For example, Rogers et
al.~\cite{Rogers2013} and Kunze et al.~\cite{Kunze2014} exploit
semantic knowledge to afford a search task. In~\cite{Rogers2013}, a
robot attaches a semantic label to each room of an environment, and
considers the semantic link between the object to search and locations
in the indoor scenario. However, the used semantic annotation is very
coarse and remains static once acquired. In~\cite{Kunze2014}, the
authors compare different areas of the environment depending on flat
surfaces and the semantic label of objects previously seen in the
scene. Also in this case the proposed framework is instantiated to a
particular task and the search is only influenced by objects
semantics. We believe that object semantics do not provide a
complete environmental knowledge and robot performance can be improved
in executing these kind of tasks by integrating information about
activities and areas where robot actions are performed.

All the aforementioned contributions exploit spatial affordances to
model a unique task and to improve robot skills in performing that
specific task. In this work, we want to introduce a general
architecture that provides the possibility to model different types of
spatial affordances simultaneously. To this end, we consider the
remarkable contribution of Lu et al.~\cite{Lu2014}. The authors
propose a layered costmap to encode different features of the
environment in order to support navigation. Their architecture enables
to formalize each layer independently, which is beneficial in the
development of robotic systems. We borrow such paradigm and propose a
modular approach in representing affordances. Additionally, we
generalize our framework by not forcing our system to only represent
navigability tasks. As shown in Section~\ref{sec:stam}, we propose a
system to semantically annotate the space of the environment in order
to support manifold high-level tasks -- of which navigability is just
an instance.

\section{STAM: Spatio-Temporal Affordance Map}
\label{sec:stam}

Affordances have been originally introduced by
Gibson~\cite{Gibson1979} as action opportunities that objects offer,
and further explored by Chemero~\cite{Chemero2003} in a more recent
work. This notion has been accordingly adopted in robotics to provide
a different perspective in representing objects and their related
actions. Here, we extend the spatial affordance theory, where the
considered ``object'' is the environment itself, by introducing the
idea of spatial semantics and spatio-temporal affordances. Spatial
semantics provides a connection between the environment and its
operational functionality -- e.g., in a surveillance task, areas that
are hidden or not entirely covered by fixed sensors present a
different ``risk semantics''.  A Spatio-Temporal Affordance (STA) is a
function that defines areas of the operational environment that afford
an action, given a particular state of the world.

\begin{figure}[t!]
  \centering
  \includegraphics[width=\textwidth]{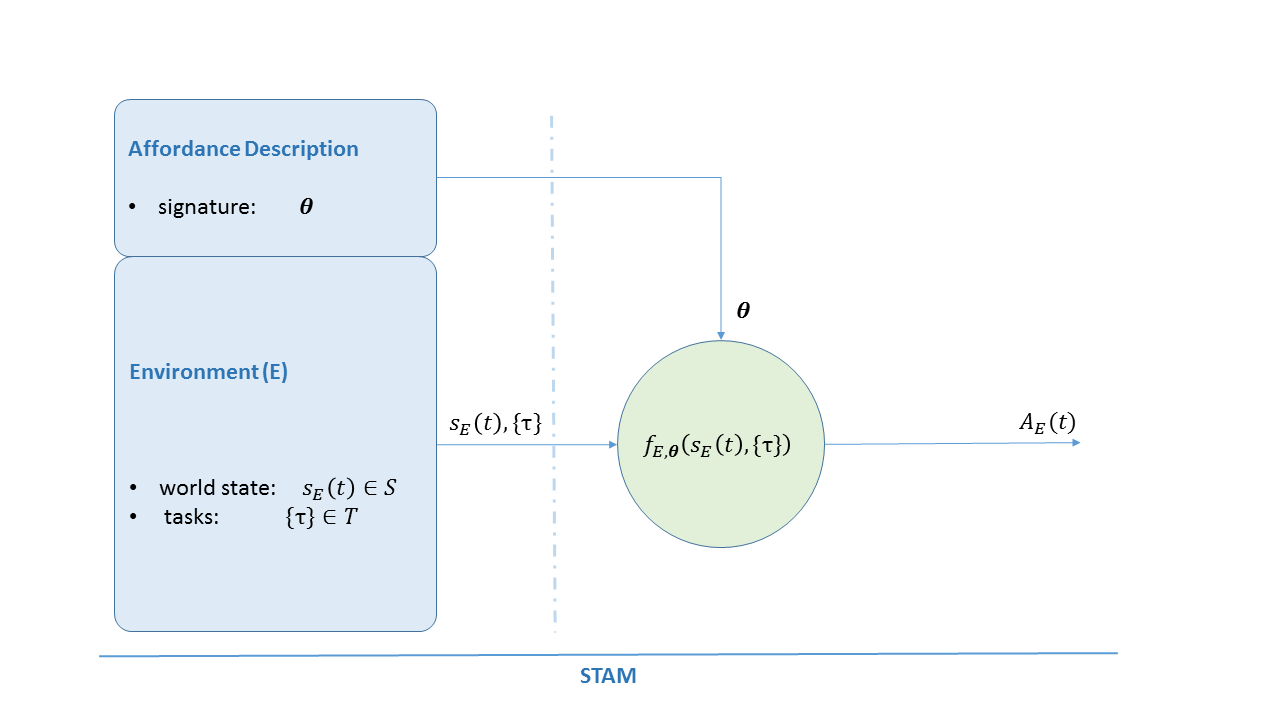}
  \caption{Spatio-Temporal Affordance Map -- STAM.}
  \label{fig:stam}
\end{figure}

\begin{defn}
  A \textit{spatio-temporal affordance (STA)} is a function

  \begin{align}
    f_{E, \boldsymbol{\theta}}: S \times T \rightarrow \mathrm{A}_E.
  \end{align}

  $f_{E, \boldsymbol{\theta}}$ depends on the environment $E$ and a
  set of parameters $\boldsymbol{\theta}$ characterizing the
  affordance function. It takes as input the state of the environment
  $s_E(t) \in S$ at time $t$, a set of tasks $\lbrace\tau(t)\rbrace
  \in T$ to be performed, and outputs a map of the environment
  $\mathrm{A}_E$ that evaluates the likelihood of each area of $E$ to
  afford $\lbrace\tau(t)\rbrace$ in $s_E$ at time $t$.
  \label{def:sta}
\end{defn}

The function $f_{E, \boldsymbol{\theta}}$ hence characterizes spatial
semantics by evaluating areas of $E$ where the set of tasks
$\lbrace\tau(t)\rbrace$ can be afforded. At each time $t$, it
generates the spatial distribution of affordances within the
environment and encodes them in a map $\mathrm{A}_E$. Then, the STA
function can be exploited by an autonomous agent as a part of a
Spatio-Temporal Affordance Map (STAM) - a representation that encodes
the semantics of the agent's actions related to the environment.

\begin{defn}
  A Spatio-Temporal Affordance Map (STAM) is a representation of the
  STA of an environment that can be (1) learned, (2) updated and (3)
  used by an autonomous agent to modify its own behavior.
  \label{def:stam}
\end{defn}

As depicted in Fig.~\ref{fig:stam}, the core element of a STAM is the
function $f_{E, \boldsymbol{\boldsymbol{\theta}}}$ introduced in
Def. \ref{def:sta}, that depends on a set of parameters
$\boldsymbol{\theta}$ obtained from an \textit{affordance description
  module} and takes as input the current state of the world and a set
of tasks from the \textit{environment module}. In particular:

\begin{itemize}
\item the affordance description module (\textbf{a-module}) is a
  knowledge base composed by a library of parameters
  $\boldsymbol{\theta}$ that characterize the STA and represent its
  \textit{signature}. The signature modifies the spatial distribution
  of affordances within the environment;

\item the \textit{environment module} (\textbf{e-module}) encodes the
  state of the world $s_E(t)$ and provides such a state to the STA
  function, by coupling it with a set of tasks $\lbrace\tau(t)\rbrace$
  to be executed in order to achieve the desired goal.
\end{itemize} 

\begin{figure}[t!]
  \centering
  \includegraphics[width=\textwidth]{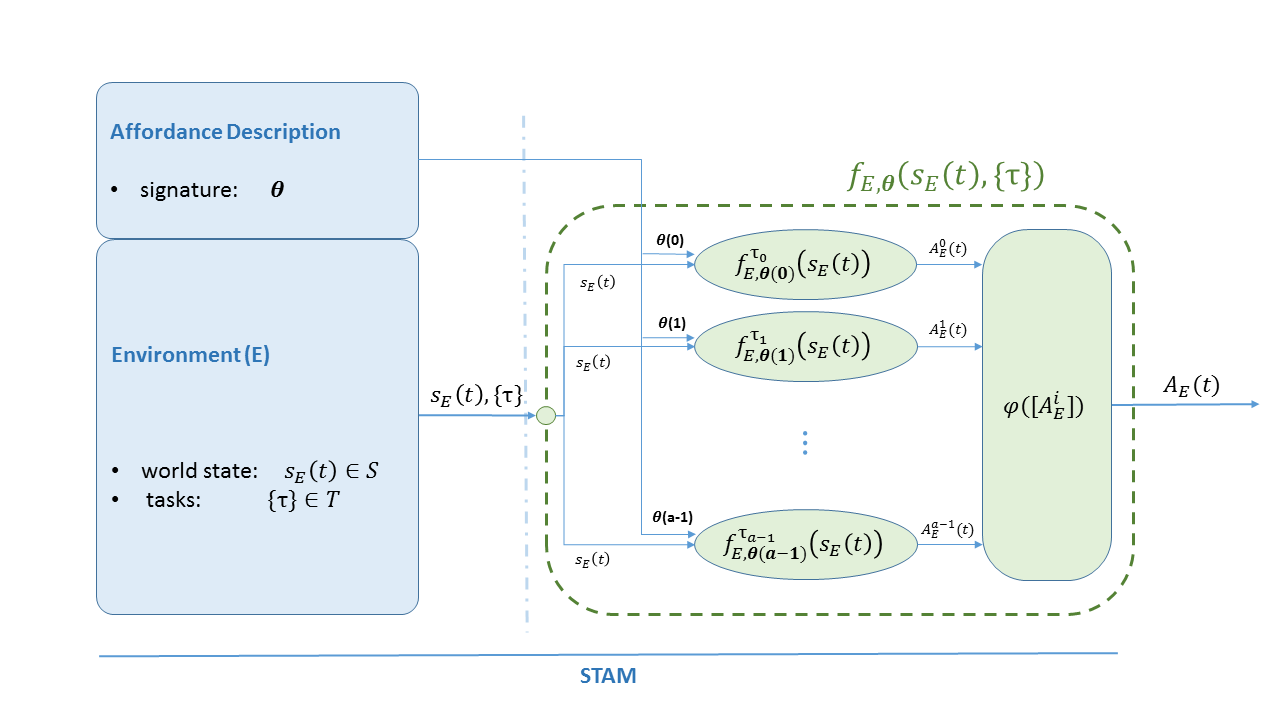}
  \caption{Spatio-Temporal Affordance Map -- STAM.}
  \label{fig:stam_multi}
\end{figure}

It is worth remarking that $f_{E, \boldsymbol{\theta}}$, $s_E$ and
$\mathrm{A}_E$ refer to a common representation of the environment $E$
that needs to be instantiated in order to enable a robot to use
STAM. Such a representation can be chosen to be a metric map, a grid
map, a topological map, or a semantic map. Additionally, STAM can be
used to interpret relations among different affordances (if there
exists) and to represent affordances individually. In fact, as shown
in Fig.~\ref{fig:stam_multi}, a STA can be seen as a composition of
different $f^{\tau_i}_{E, \boldsymbol{\theta}(i)}$ functions ($i \in
[0, a-1]$, where \textit{a} is the number of affordances), each
modeling the spatial distribution $\mathrm{A}_E^i$ of a particular
affordance in $E$. These are then combined by a function $\phi$, that
takes as input all the $\mathrm{A}_E^i$ and outputs a map
$\mathrm{A}_E$ that satisfies $\lbrace\tau(t)\rbrace$, according to
the considered affordances.

\subsection{Generating a Spatio-Temporal Affordance Map}
\label{sec:generating_stam}

The affordance map $\mathrm{A}_E$ is a representation of the
operational environment that evaluates $E$ with respect to the current
state of the world and encodes areas of $E$ where a particular task
can be afforded. For instance, in the case in which the environment is
represented as a grid-map, $\mathrm{A}_E$ encodes in each cell the
likelihood of a given area to afford an action. According to
Def.~\ref{def:sta}, the generation of $\mathrm{A}_E$ directly depends
on a general set of parameters $\boldsymbol{\theta}$ -- the affordance signature --
that modify how affordances model the space. Hence, they constitute
the main vehicle to shape affordances and need to be carefully
designed or learned. In the first case, accurate understanding of each
parameter $\boldsymbol{\theta}$ and the function $f_{E, \boldsymbol{\theta}}$ is required. In
the latter case, the STA function can be implemented as regression or
classification algorithm, and standard gradient-based methods can be
used to update $\boldsymbol{\theta}$.  For instance, when learning affordances from
observations of other agents' behaviors (e.g., humans) a neural
network could be used. In this case, the set of parameters $\boldsymbol{\theta}$
would represent the connection weights between different layers and
they could be computed by means of back-propagation.

\section{Using STAM on a Robot}
\label{sec:stam_robot}

STAM is intended to directly influence the behavior of an autonomous
agent and, in particular, the navigation stack of a mobile robot. We
consider the case in which the robot navigation system relies upon
standard costmap-based techniques~\cite{Lu2014}. In contrast to
previous work in this field, we are not interested in enabling a robot
to ``\textit{go from point A to point B}'', but we aim at making the
agent capable to ``\textit{go from A to} $\beta$", where $\beta \in
B'$ is a set of ``good'' poses obtained from the map $\mathrm{A}_E$
generated by STAM. Such poses intrinsically respect spatio-temporal
constrains imposed by the considered affordances. Among these, the
selection of the final pose can be based upon different criteria, such
as the top scoring area in $\mathrm{A}_E$, the nearest area to the
robot, the biggest area, or a combination of these
criteria. Nevertheless, we also want the robot to decide how to
navigate the environment by selecting the path accordingly to the
affordances imposed by the task. To this end, we can directly use
$\mathrm{A}_E$ to effectively crop out all the trajectories of the
robot that cross areas violating affordance constraints. In
particular, we can substitute the costmap with a \textit{gainmap} that
encodes high-level information extracted from STAM. Accordingly, the
robot will not follow the cheapest path, as in ``usual'' costmap-based
systems, but it will maximize its gain over the generated gainmap.
Such a map is generated as a function of the normalized cost map and
the likelihood obtained from $\mathrm{A}_E$.

\begin{align}
  \label{eq:gainmap}
  m(cost, likelihood, \lambda) &= \lambda (1 - cost) + (1- \lambda) likelihood,
\end{align}

with $\lambda \in [0,1]$. In this respect, we are modifying the
navigation systems of an autonomous robot by transferring high-level
information encoded in $\mathrm{A}_E$ into the navigation system.

\section{Experiments}
\label{sec:experiments}
In order to evaluate of our approach we perform an analysis of the
learned affordance model. To this end, we exploit expert
demonstrations to teach a robot how to correctly interpret the
environment when performing a following task.  Then, we evaluate the
learned model by reporting the affordance map $\mathrm{A}_E$ generated
by the affordance function and the prediction error of the regression
algorithm after each demonstration.

\subsection{Affordance of a Following Task}
\label{sec:case_study}

We consider a robot that has to perform a following task. In this
case, the areas of the environment $E$ that afford the task depend on
manifold factors such as general rules (e.g., forbidden areas), user
preferences (that can be encoded in the set of parameters
$\boldsymbol{\theta}$) and the position of the followed person
(encoded in the state of the environment $s_E$). According to
Def.~\ref{def:sta}, we can generate $\mathrm{A}_E$ and identify robot
poses that support the execution of the task. To this end, we encode
the pose ${\langle x_T, y_T, \alpha_T \rangle}$ of the target $T$ to
follow in the state $s_E(t)$. Additionally, we use Gaussian Mixture
Models (GMMs) and Gaussian Mixture Regression to represent and
implement the function $f_{E, \boldsymbol{\theta}}$. The signature
$\boldsymbol{\theta}$ of the STA function is hence composed as a tuple
$\boldsymbol{\theta} = \langle \pi_1, \mu_1, \Sigma_1, \dots, \pi_N,
\mu_N, \Sigma_N \rangle$, where $\pi_i$ is the prior, $\mu_i$ the mean
vector and $\Sigma_i$ the covariance matrix of a mixture of $N$
Gaussians.

\begin{figure}[t!]
  \centering
  \includegraphics[trim = 30mm 5mm 30mm 5mm, clip,width=\textwidth]{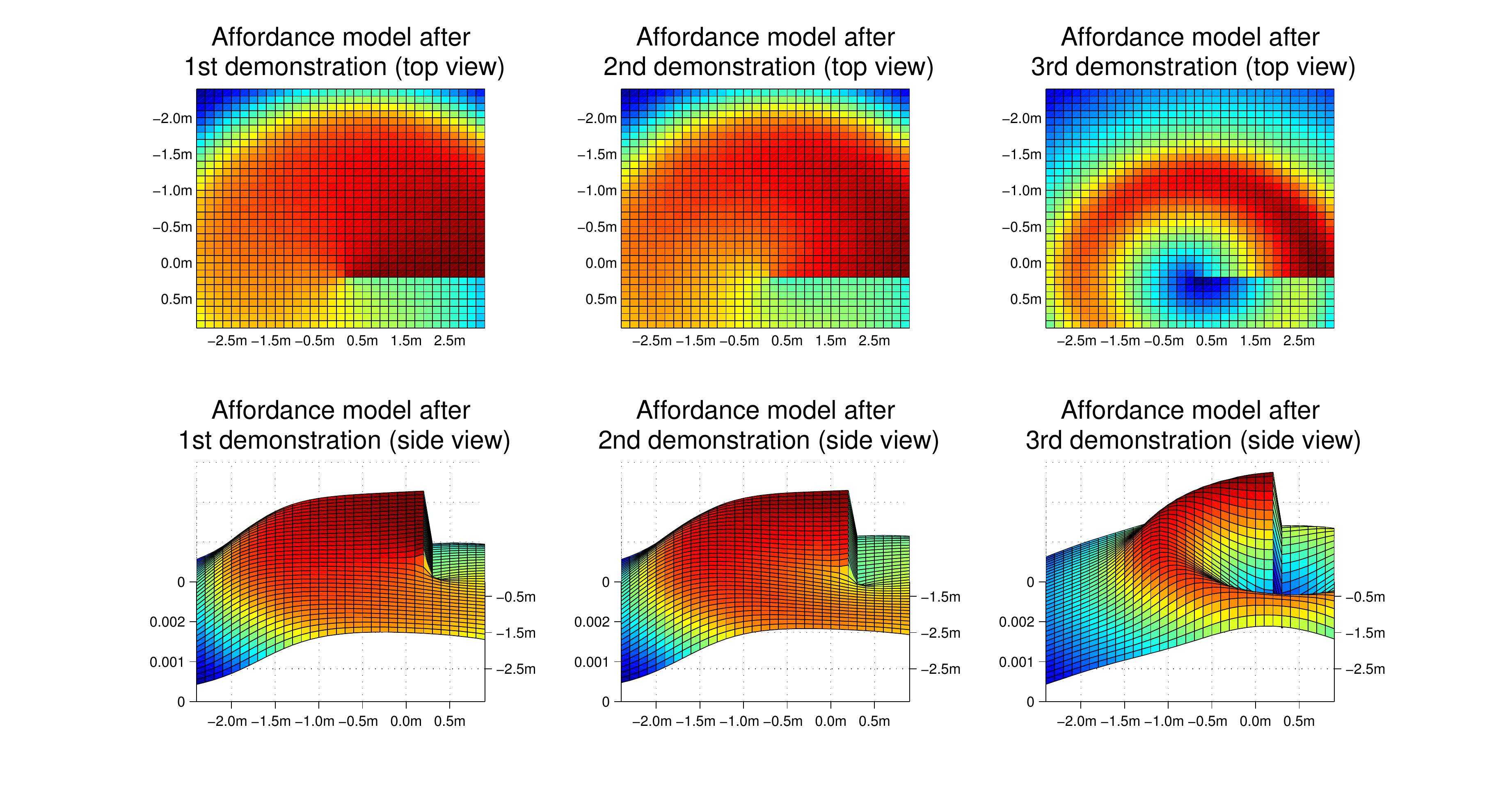}
  \caption{Spatio-temporal affordance of a following task learned with
    increasing number of expert demonstrations. Here, the target is
    located at the origin and the plots represent the probability
    density function of a pose to afford the task. The plots, whose
    coordinates are expressed in meters, show that the model is able
    to represent both minimum and maximum distances from the target,
    in accordance with the data provided as demonstrations.}
  \label{fig:example_to_follow}
\end{figure}

In this experiment, the signature $\boldsymbol{\theta}$ is learned from
demonstration of different experts. To collect expert data we setup
two robots in a simulated environment -- one randomly navigates, the
other is controlled by an expert through a joystick and follows the
target robot $T$ by always moving between a minimum and maximum
distance from it. During these sessions, the state $s_E(t)$, as
defined above, is recorded at each time instant together with the pose
${\langle x_F, y_F, \alpha_F \rangle}$ of the follower $F$. The
collected measurements are provided as input to the GMM and, by using
Expectation Maximization, the tuple $\boldsymbol{\theta} = \langle \pi_1,
\mu_1, \Sigma_1, \dots, \pi_N, \mu_N, \Sigma_N \rangle$ that best fits
the data is determined. In our experiments, prior to Expectation
Maximization, the model has been initialized with k-means and a set of
candidate GMMs has been computed with up to 8 components; the number
of components has then been selected to minimize the Bayesian
Information Criterion.

The learned model is used by the follower to determine, through
Gaussian Mixture Regression, areas of $E$ that enable the robot to
execute the task and, hence, to generate $\mathrm{A}_E$. In
particular, the output of the regression consists of a mean vector and
covariance matrix that enable us to infer the probability distribution
(shown in \figurename~\ref{fig:example_to_follow}) of the follower
pose, given the target pose for the following task $\tau$. In this
example, no specific constraint is imposed to the robot for the
selection of its path. Hence, the agent can select the pose that
maximizes its profits over the gainmap computed according to
Eq.~\ref{eq:gainmap}, and reach it by following the shortest path.

\begin{figure}[t]
  \centering
  \begin{varwidth}{0.5\linewidth}
    \subfigure[]{\includegraphics[trim = 20mm 5mm 20mm 5mm, clip,width=0.9\textwidth]{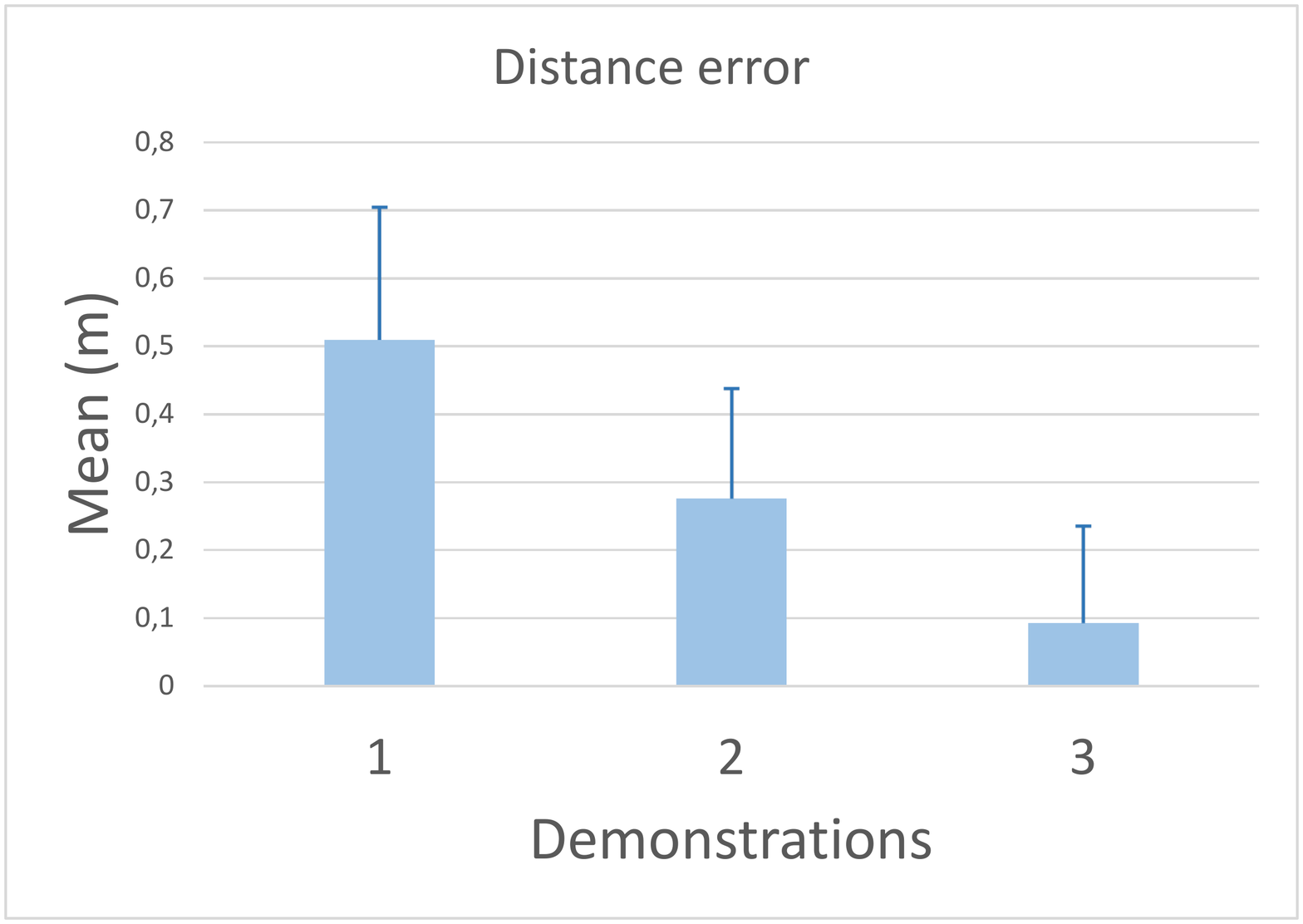}}
  \end{varwidth}
  \begin{varwidth}{0.5\linewidth}
    \subfigure[]{\includegraphics[trim = 20mm 5mm 20mm 5mm, clip,width=0.9\textwidth]{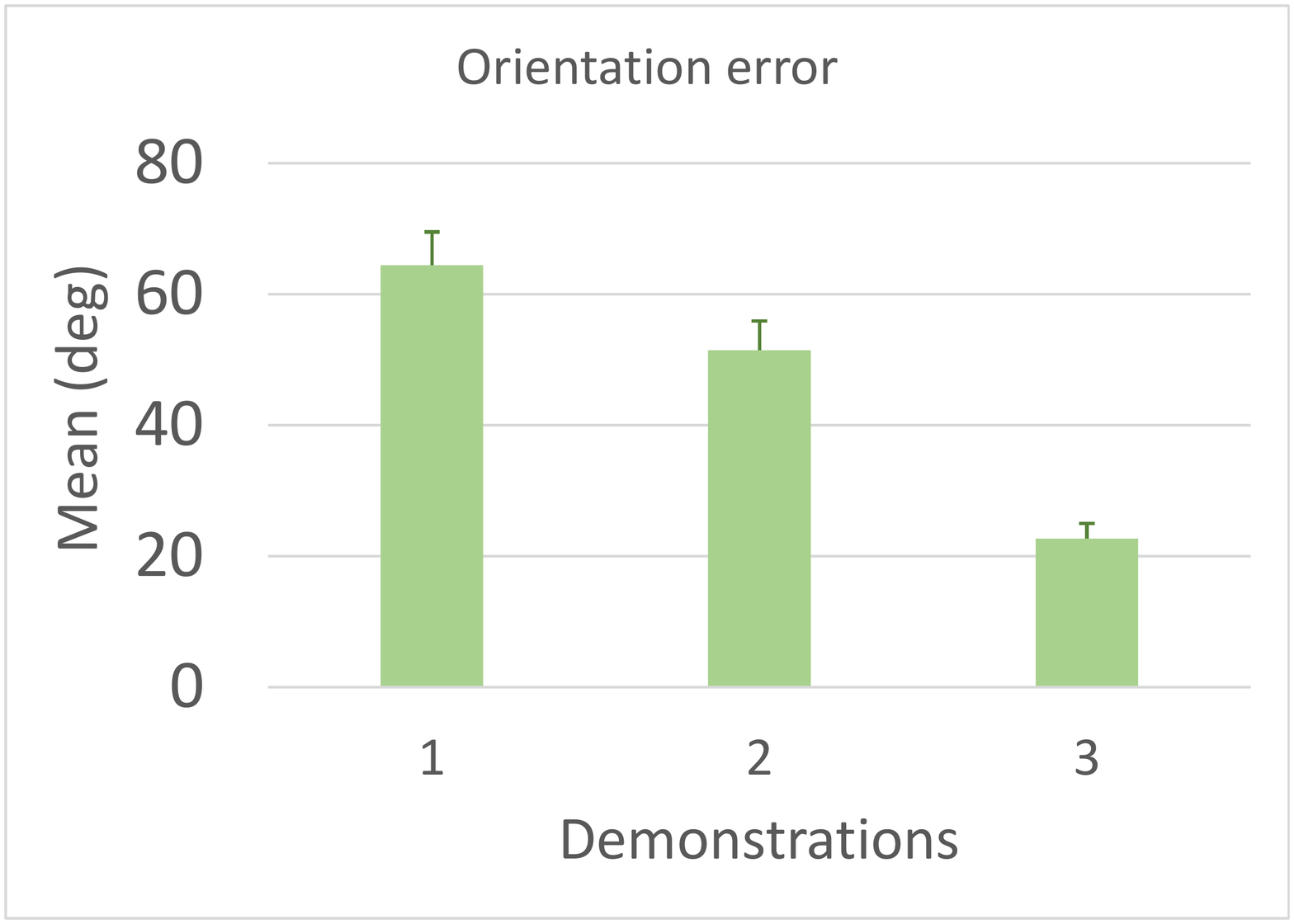}}
  \end{varwidth}
  \caption{Error of the best pose, selected according to the learned
    model, against the expert behavior. On the left we report (a) the
    mean and standard deviation of relative distance error between the
    follower and the target, while on the right (b) the mean and
    standard deviation of the relative orientation error are shown.
    These values have been obtained by running 20 experiments and
    incrementally using three expert demonstrations (arranged on the
    x-axis).}
  \label{fig:error_analysis}
\end{figure}

Finally, we report an analysis of the prediction error of the
affordance model generated by the regression algorithm. To this end,
we use expert data collected in three different demonstrations in an
incremental fashion -- after each demonstration we append new training
examples to the previous dataset. Then, we generate the affordance
model by splitting the dataset into two distinct parts. One is used to
learn the affordance model, while the other is used to compute the
error of the best pose, selected according to the learned model,
against the expert behavior (the ground-truth). To evaluate our model,
we ran the experiment 20 times. Accordingly,
\figurename~\ref{fig:error_analysis} shows the mean and standard
deviation of the prediction errors of the relative distance (a) and
orientation (b) between the target and the follower position. It is
worth remarking that, as soon as the affordance model becomes more
accurate (\figurename~\ref{fig:example_to_follow}), the prediction
errors of both the distance and orientation decay.

\section{Conclusion}
\label{sec:conclusion}

In this paper we presented and formalized Spatio-Temporal Affordances
(STA) and Spatio-Temporal Affordance Maps (STAM) as a novel framework
to represent spatial semantics. This is a relevant problem since, by
providing a connection between the environment and its operational
functionality, spatial semantics leads to a proper interpretation of
the environment and hence to a better execution of robot tasks. To
test this representation, we implemented STAM and learned the
affordance model of a following task by exploiting expert
demonstrations. Specifically, we set up a simulated environment where
human experts could teach the robot how to correctly interpret the
environment when performing a following task. After training, we let
our system infer the best position to be in order to follow a
target. Results show that (1) the mapping between the space and its
affordance is qualitatively valid and (2) the error generated by the
use of our model decreases when it becomes more accurate, through the
use of a larger number of expert demonstrations.

Nevertheless, learning the affordance of a following task is only a
simple and specific use case of STAM. For this reason, in future work
we aim at using STAM to run different experiments with manifold tasks
and, specifically, to enable a robot to interpret spatial semantics to
improve human-robot interactions.

\bibliographystyle{splncs03}
\bibliography{refs}

\begin{thebibliography}{10}
\providecommand{\url}[1]{\texttt{#1}}
\providecommand{\urlprefix}{URL }

\bibitem{Chemero2003}
Chemero, A.: An outline of a theory of affordances. Ecological Psychology
  15(2),  181--195 (2003)

\bibitem{Diego2011}
Diego, G., Arras, T.K.O.: Please do not disturb! minimum interference coverage
  for social robots. In: 2011 IEEE/RSJ International Conference on Intelligent
  Robots and Systems. pp. 1968--1973 (Sept 2011)

\bibitem{Epstein2015}
Epstein, S.L., Aroor, A., Evanusa, M., Sklar, E., Parsons, S.: Navigation with
  learned spatial affordances. In: COGSCI (2015)

\bibitem{Gibson1979}
Gibson, J.J.: The ecological approach to visual perception. Houghton Mifflin,
  Boston (1979)

\bibitem{Kapadia2009}
Kapadia, M., Singh, S., Hewlett, W., Faloutsos, P.: Egocentric affordance
  fields in pedestrian steering. In: Proceedings of the 2009 Symposium on
  Interactive 3D Graphics and Games. pp. 215--223. I3D '09, ACM, New York, NY,
  USA (2009)

\bibitem{Kim2015}
Kim, D.I., Sukhatme, G.S.: Interactive affordance map building for a robotic
  task. In: Intelligent Robots and Systems (IROS), 2015 IEEE/RSJ International
  Conference on. pp. 4581--4586. IEEE (2015)

\bibitem{Koppula2013}
Koppula, H.S., Gupta, R., Saxena, A.: Learning human activities and object
  affordances from rgb-d videos. The International Journal of Robotics Research
   32(8),  951--970 (2013)

\bibitem{Kunze2014}
Kunze, L., Burbridge, C., Hawes, N.: Bootstrapping probabilistic models of
  qualitative spatial relations for active visual object search. In: AAAI
  Spring Symposium 2014 on Qualitative Representations for Robots. Stanford
  University in Palo Alto, California, US (March, 24--26 2014)

\bibitem{Lu2014}
Lu, D.V., Hershberger, D., Smart, W.D.: Layered costmaps for context-sensitive
  navigation. In: 2014 IEEE/RSJ International Conference on Intelligent Robots
  and Systems. pp. 709--715 (Sept 2014)

\bibitem{Luber2011}
Luber, M., Tipaldi, G.D., Arras, K.O.: Place-dependent people tracking.
  Int.~Journal of Robotics Research  30(3),  280--293 (2011)

\bibitem{Montesano2008}
Montesano, L., Lopes, M., Bernardino, A., Santos-Victor, J.: Learning object
  affordances: From sensory--motor coordination to imitation. IEEE Transactions
  on Robotics  24(1),  15--26 (Feb 2008)

\bibitem{Pandey2012}
Pandey, A.K., Alami, R.: Taskability graph: Towards analyzing effort based
  agent-agent affordances. In: RO-MAN, 2012 IEEE. pp. 791--796. IEEE (2012)

\bibitem{Rogers2013}
Rogers, J.G., Christensen, H.I.: Robot planning with a semantic map. In:
  Robotics and Automation (ICRA), 2013 IEEE International Conference on. pp.
  2239--2244 (May 2013)

\end{thebibliography}

\end{document}